\pdfoutput=1

\documentclass[11pt]{article}

\usepackage[preprint]{acl}

\usepackage{times}
\usepackage{latexsym}

\usepackage[T1]{fontenc}

\usepackage[utf8]{inputenc}

\usepackage{microtype}

\usepackage{inconsolata}

\usepackage{amsfonts}
\usepackage{amsthm}
\usepackage{amsmath}
\usepackage{amssymb}
\usepackage{multirow}
\usepackage{booktabs}
\usepackage{colortbl}
\usepackage{threeparttable}
\usepackage{array}
\usepackage{graphicx}
\usepackage{svg}
\usepackage{subcaption}
\usepackage{longtable}
\usepackage{wrapfig}
\usepackage{xcolor}  
\newtheorem{proposition}{Proposition}

\usepackage{hyperref}
\usepackage{float}
\definecolor{customblue}{HTML}{DAE8FC}
\definecolor{customgreen}{HTML}{D6E8D4}
\definecolor{customorange}{HTML}{FFE6CD}
\definecolor{custompurple}{HTML}{E2D6E8}

\title{Direct Value Optimization: \\Improving Chain-of-Thought Reasoning in LLMs with Refined Values}

\author{
    Hongbo Zhang\textsuperscript{\rm 1,2,\footnotemark[1]},  
    Han Cui\textsuperscript{\rm 2,\footnotemark[1]},
    Guangsheng Bao\textsuperscript{\rm 1,2,\footnotemark[1]},
    Linyi Yang\textsuperscript{\rm 4}, Jun Wang\textsuperscript{\rm 4}, 
    Yue Zhang\textsuperscript{\rm 2,3,\footnotemark[2]}
    \\
    \textsuperscript{1} Zhejiang University \\
    \textsuperscript{2} School of Engineering, Westlake University \\
    \textsuperscript{3} Institute of Advanced Technology, Westlake Institute for Advanced Study \\
    \textsuperscript{4} University College London \\
    \texttt{\{zhanghongbo,cuihan,baoguangsheng,zhangyue\}@westlake.edu.cn} \\
    \texttt{
    yanglinyiucd@gmail.com\quad jun.wang@cs.ucl.ac.uk}\\
}

\begin{document}
\maketitle

\renewcommand{\thefootnote}{\fnsymbol{footnote}}
\footnotetext[1]{Equal contribution. \footnotemark[2]Corresponding author.}
\renewcommand{\thefootnote}{\arabic{footnote}}

\begin{abstract}
We introduce Direct Value Optimization (DVO), an innovative reinforcement learning framework for enhancing large language models in complex reasoning tasks. Unlike traditional methods relying on preference labels, DVO utilizes value signals at individual reasoning steps, optimizing models via a mean squared error loss. The key benefit of DVO lies in its fine-grained supervision, circumventing the need for labor-intensive human annotations. Target values within the DVO are estimated using either Monte Carlo Tree Search or an outcome value model. Our empirical analysis on both mathematical and commonsense reasoning tasks shows that DVO consistently outperforms existing offline preference optimization techniques, even with fewer training steps. These findings underscore the importance of value signals in advancing reasoning capabilities and highlight DVO as a superior methodology under scenarios lacking explicit human preference information.

\end{abstract}

\section{Introduction}
Large language models (LLMs) have demonstrated exceptional performance across a wide range of natural language processing (NLP) tasks \citep{yu2023natural,azerbayev2023llemma,reid2024gemini,phoenix-2023,achiam2023gpt,yang2024qwen2}. However, when addressing complex problems such as mathematical reasoning \citep{yue2023mammoth,yu2023metamath}, logical reasoning \citep{liu2023evaluating,teng2023glore}, and multi-step planning problems \citep{Hao2023ReasoningWL}, they can fall into incorrect reasoning paths due to errors in intermediate reasoning steps~\citep{cobbe2021training,shen2021generate,bao2025llms}. 

To address this issue, recent studies leverage reinforcement learning (RL) to optimize the reasoning process according to feedback derived from golden answers ~\citep{shao2024deepseekmath,luo2023wizardmath,chen2024huatuogpt,kazemnejad2024vineppo}, significantly enhancing the robustness. Among these, converting reward signals into preference labels has been a standard practice, where multiple responses are generated for one question and the optimal response reaching at the correct answer is labeled as the preferred response in contrast to others. Preference labels are then used to train models through a cross-entropy loss or a contrastive alternatives. For example, DPO \citep{rafailov2023directpreferenceoptimizationlanguage}, KTO \citep{ethayarajh2024kto}, and RFT~\citep{yuan2024advancing,yuan2023scaling} optimize language models based on preferences at the token level, while recent process supervision \cite{lightman2023letsverifystepstep}, CPO~\citep{zhang2024chain}, Step-DPO~\citep{lai2024step}, SVPO \citep{chen2024step} and Iterative Preference Learning \citep{xie2024monte} generate preferences and optimize LLMs at the step level. 

Preference labels as training targets are simple and convenient for implementation. 
However, rewards from preference labels lack fine-grained signals and can fail to preserve critical value information~\cite{liu2023making,chen2024step}. 

A key limitation of preference-based optimization is its reliance on pairwise comparisons, which only indicate relative preference between two choices rather than providing an absolute ranking of multiple steps. In contrast, stepwise value signals offer a continuous ranking, allowing the model to make more informed decisions beyond binary comparisons. By leveraging value signals, models can better understand reasoning dynamics, prioritizing steps that enhance clarity, reduce ambiguity, and improve multi-step reasoning efficiency.
\begin{figure*}[t]
    \centering\small
    \setlength{\abovecaptionskip}{6pt}  
    \includegraphics[width=1.0\textwidth]{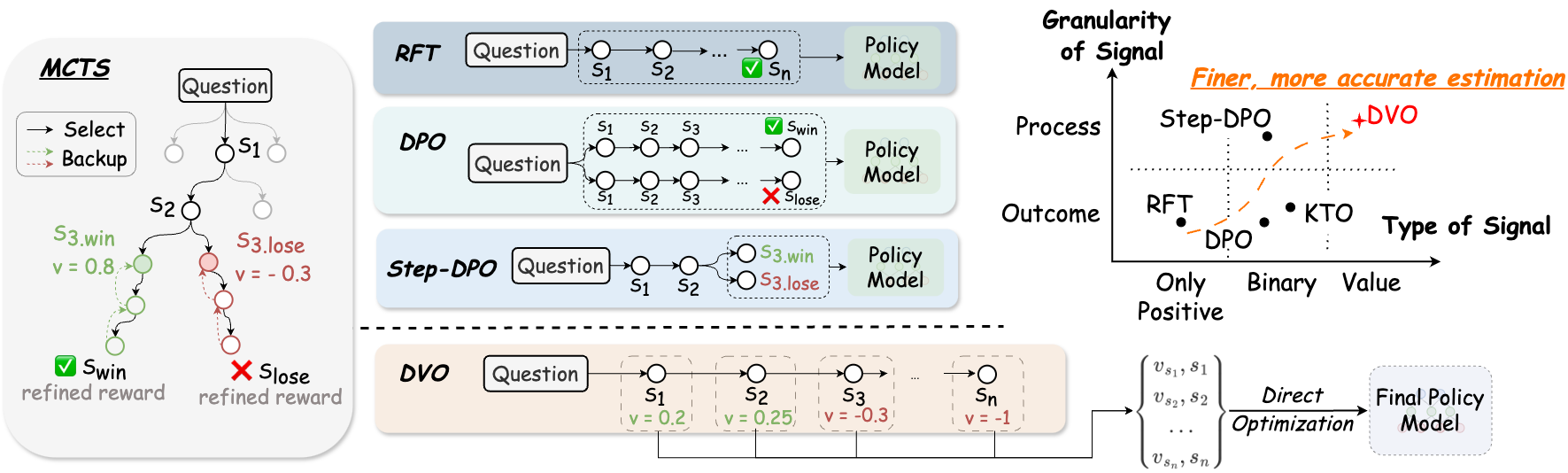}
    \caption{Overview of Direct Value Optimization. Compared with other self-improvement methods, DVO stands out by using MCTS to generate self-explored data and directly aligning the policy model with value estimations. This provides an efficient self-improvement framework that finely tunes the model to maximize total expected rewards.
    }
    \label{fig:dvo_framework}
    \vspace{-5pt}
\end{figure*}

To address this issue, we propose \emph{Direct Value Optimization (DVO)}, a simple yet effective reinforcement learning (RL) framework for large language models (LLMs). As shown in Fig.~\ref{fig:dvo_framework}, DVO estimates target values at each reasoning step and aligns the model with these values using a mean squared error (MSE) loss. 
To make use of widely available outcome-supervised data, we infer stepwise value signals from the final outcome, providing process-level guidance for RL. The key distinction between DVO and other offline RL methods lies in its ability to optimize LLMs using value signals without requiring process-level human labeling, thereby eliminating the need for labor-intensive annotations while retaining fine-grained supervision. In DVO, target values can be estimated using various methods. 
Specifically, we explore two primary approaches: monte carlo tree search (MCTS) \citep{kocsis2006bandit,coulom2006efficient} and outcome value model \citep{yu2023outcome,feng2023alphazero}.

We conduct extensive experiments on both mathematical reasoning tasks and commonsense reasoning tasks, evaluating models with various sizes. Results show that DVO outperforms other offline preference optimization algorithms across multiple benchmarks under the same number of training steps. For example, a three rounds of DVO training improves Llama3-8B-Instruct’s~\citep{dubey2024llama} accuracy on GSM8K~\citep{cobbe2021training} from $74.6\%$ to $80.6\%$, on MATH~\citep{hendrycks2021measuring} from $22.5\%$ to $26.5\%$, and on AGIEval-Math~\citep{zhong2023agieval} from $23.5\%$ to $27.9\%$. Finally, comprehensive analytical experiments highlight the critical role of value signals in reasoning tasks. Our code is available at \url{https://github.com/StevenZHB/DVO}.
\begin{figure*}[t]
    \centering\small
    \setlength{\abovecaptionskip}{6pt}  
    \includegraphics[width=1.0\textwidth]{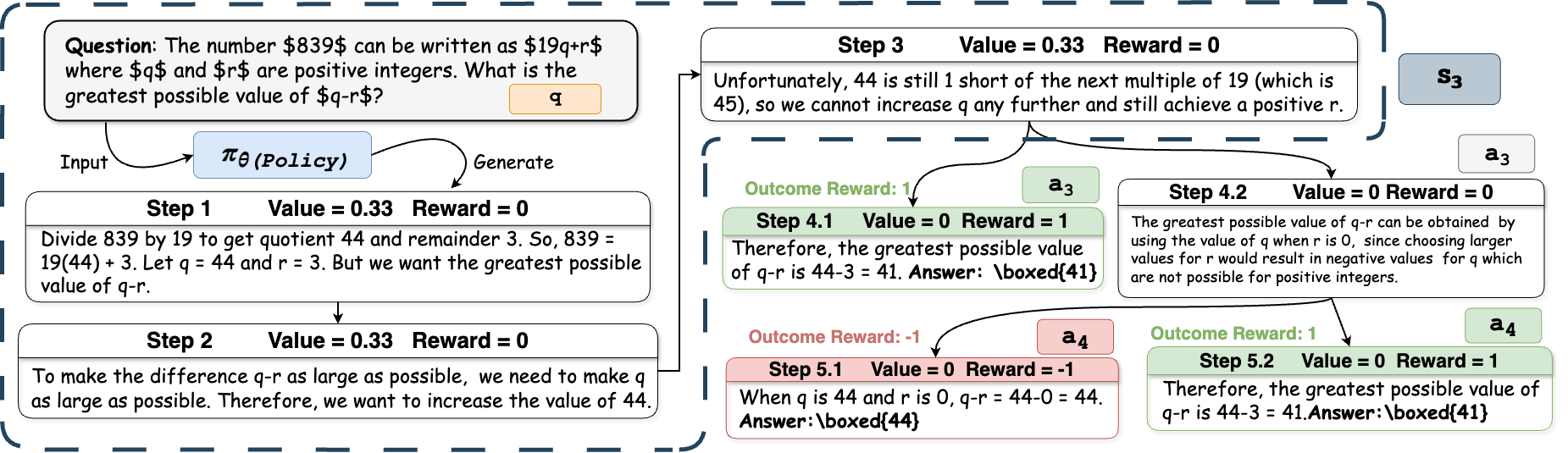}
    \caption{Illustration of Step-by-step reasoning process, each node represents a step with its corresponding reward and value estimation.}
    \vspace{-5pt}
    \label{fig:step_example}
\end{figure*}

\section{Related Work}

\textbf{Improving Reasoning in LLMs.} Existing studies improve the robustness of LLM reasoning at various stages. Some focus on the pre-training phase by curating pretraining datasets to enhance mathematical knowledge \citep{shao2024deepseekmath,azerbayev2023llemma}, some address it during instruction fine-tuning using high-quality reasoning QA datasets \citep{mitra2023orca,mukherjee2023orca,yue2023mammoth,yu2023metamath}, while others tackle the reasoning challenge at the reinforcement learning stage by using human or AI feedback for optimization \citep{yuan2024advancing,lai2024step}. Additionally, several studies improve reasoning at the inference stage through prompting techniques that search for better reasoning paths in trees or graphs \citep{yao2024tree,besta2024graph,wang2023plan,yang2024buffer,zheng2023take}. Our work aligns most closely with methods that focus on the reinforcement learning stage but differs by improving LLMs by using value signals. 

\textbf{Reinforcement Learning in LLMs.} Reinforcement learning is widely used in LLM post-training to learn implicit rewards from human feedback~\citep{ouyang2022training,dubey2024llama,yang2024qwen2}. The standard pipeline trains a reward model and updates the policy via PPO~\citep{schulman2017proximal}. DPO~\citep{rafailov2023directpreferenceoptimizationlanguage} shows that the log ratio between optimal policy and reference models expresses the corresponding reward model, allowing direct policy updates based on the preference data. \citet{rafailov2024r} further derived the DPO objective, validating its ability to learn token-level rewards, while subsequent works refine DPO for practical applications~\citep{ethayarajh2024kto,chen2024noise,azar2024general}. Our approach extends DPO by aligning the policy model with value estimation, eliminating the need for pairwise data and leveraging value signals for enhanced reasoning. While recent work like DQO~\citep{liu2024enhancing} and OREO~\cite{wang2024offline} also aims to improve LLM reasoning within the value-based framework, they differ from our approach in that they require training a separate value model alongside the policy model. In contrast, our method is more closely aligned with DPO, requiring only the policy model and a reference model during training.

\textbf{Tree-search Guided Reasoning in LLMs.} Tree search has recently emerged as a key technique in LLM reasoning. Prior work has used prompt engineering to estimate expected returns for reasoning steps, guiding models toward optimal answers~\citep{yao2024tree,feng2023alphazero,besta2024graph}. Subsequent studies employed tree search to generate training data for RL or fine-tuning. \citet{xie2024monte} and \citet{chen2024step} use value estimates from tree search to create step-level preference pairs, while \citet{feng2023alphazero} and \citet{chen2024alphamath} train value models from these estimates for guided decoding. Our work also leverages MCTS, but unlike \citet{feng2023alphazero} and \citet{chen2024alphamath} who focus on improving the inference stage, we directly train the policy model using MCTS value estimates. Additionally, compared to \citet{xie2024monte} and \citet{chen2024step}, we make use of value estimates instead of preference, obtaining stronger results.

\section{Baseline: Direct Preference Optimization}
Direct Preference Optimization (DPO)~\citep{rafailov2023directpreferenceoptimizationlanguage} was designed to align LLMs with human preference labels. Recent work~\citep{yuan2023scaling, ethayarajh2024kto, chen2024noise} has extended DPO to optimize LLMs with reward signals by converting them into preference labels through specific algorithms. While DPO was initially developed for the contextual bandit setting, it can be validated within token-level MDP~\cite{rafailov2024r}. In the RLHF pipeline, we first train a reward model based on the Bradley-Terry framework~\citep{bradley1952rank}, which is well-suited for modeling implicit human preferences:
\begin{equation}
\small
    P_{\text{BT}}(y_1 \succ y_2 \mid x) = \frac{\exp(r(x, y_1))}{\exp(r(x, y_1)) + \exp(r(x, y_2))},
\end{equation}
where $x$ is the prompt, $y_1$, $y_2$ are two completions. The optimal policy, $Q$-function, and $V$-function learned from a reward function $R$ are denoted as $\pi^*$, $Q^*$, and $V^*$, respectively. satisfying the relation in Eq.~\eqref{eq:optimal_relation}.
Using $\pi^*$, $Q^*$, and $V^*$, the accumulated step-level reward is:
\begin{align*}
\small
R(x,y) &= \sum_{t=1}^{T-1}(Q^*(\mathbf s_t,\mathbf a_t)-V^*(\mathbf s_{t+1}))\\
&=V^*(\mathbf s_0) - V^*(\mathbf s_T) + \beta\log\pi^*(y\mid x)\\
&= V^*(\mathbf s_0) + \beta\log\pi^*(y\mid x)
\end{align*}
where we assume $V^*(s_T) = 0$. To prevent the model from deviating excessively from its prior behavior during training, the reward function is: 
\begin{equation*}
\small
    R(a_t \mid s_t) = \beta \log \pi_\text{ref}(a_t \mid s_t) + r(a_t \mid s_t),
\end{equation*}
which leads to the reward model as:
\begin{equation*}
\small
     r(x, y) = \beta \log \frac{\pi^*(y \mid x)}{\pi_\text{ref}(y \mid x)} + V^*(\mathbf s_0).
\end{equation*}

This establishes a connection between the learned policy and the reward model, enabling direct optimization of the policy without the need for explicit reward modeling. Substituting the regularized reward into the Bradley-Terry framework yields the DPO objective:

\begin{equation}
\small
\begin{split}
\mathcal{L}_{\text{DPO}} = -\mathbb{E}_{(x,y_w, y_l)} \bigg[ 
& \log\sigma\bigg(  \beta \log \frac{\pi_\theta(y_w \mid x)}{\pi_{\text{ref}}(y_w \mid x)} \\
& - \beta \log \frac{\pi_{\theta}(y_l \mid x)}{\pi_{\text{ref}}(y_l \mid x)} 
\bigg) \bigg],\\
\label{eq:dpo_objective}
\end{split}
\end{equation}
where $\sigma$ denotes the sigmoid function, and $y_w$, $y_l$ represent winning and losing responses respectively. The baseline term $V^{*}(\mathbf s_0)$ cancels out in pairwise comparisons, eliminating the need for explicit value estimation ($\mathbf s_0$ is the same for both $y_w$ and $y_l$). This derivation is based on a token-level MDP but can also be applied to the step-level MDP introduced later in this work.

\section{Direct Value Optimization}
We derive a framework that optimize reasoning of LLMs using trajectories along with their estimated stepwise values by extending soft $Q$-learning. To our knowledge, this is the first work to directly optimize LLMs using estimated value signals within a $Q$-learning framework.

\subsection{Problem Definition: A Stepwise MDP}
As Fig.~\ref{fig:step_example} illustrates, we formulate the reasoning problem as a stepwise Markov Decision Process (step MDP), where a whole reasoning process is split into several small but coherent reasoning steps, each representing a link in the chain of thoughts.

Formally, at each step $t$, the policy model is in a state $\mathbf s_t$, which consists of the current question $q$ (sampled from a distribution $\mathcal{D}$)  and a sequence of prior steps $[y_1, y_2, \dots, y_{t-1}]$, denoted as $\mathbf s_t = \{q, y^{<t}\}$. The policy model $\pi_{\theta}$ selects the next action $\mathbf a_t$ from the set of possible steps according to the probability distribution $\pi_\theta(\mathbf a_t \mid \mathbf s_t)$, guiding the reasoning process forward. The goal of the model is to produce a sequence of steps that ultimately leads to the final answer. 

To encourage the model to perform robust reasoning steps that lead to the correct outcome, we define a reward function $r(\mathbf s_t,\mathbf a_t)$. The reward evaluates the quality of each reasoning step, ensuring that the model focuses on generating logical and accurate transitions. In this context, the reasoning process is deterministic: the next state $\mathbf{s}_{t+1}$ is fully determined by the current state $\mathbf{s}_t$ and the selected action $\mathbf{a}_t$. As a result, we omit explicit references to transition probabilities $P(\mathbf{s}_{t+1} \mid \mathbf{s}_t, \mathbf{a}_t)$ in subsequent equations for simplicity.
\label{sec:step_MDP_definition}
\subsection{LLM as Soft $Q$-Function}
Maximum entropy reinforcement learning introduces an entropy term to encourage exploration, balancing reward maximization and stochasticity for robust decision-making \citep{rafailov2024r,ziebart2010modeling,haarnoja2017reinforcement}. In this framework, the Soft $Q$-function and Soft $V$-function are defined as follows:
\begin{equation}
\small
\begin{aligned}
Q_{\text{soft}}(\mathbf s_t, \mathbf a_t) &= R(\mathbf s_t, \mathbf a_t) + V_{\text{soft}}(\mathbf s_{t+1}), \\
V_{\text{soft}}(\mathbf s_t) &= \beta \log \int_{\mathcal{A}} \exp(\frac{1}{\beta} Q_{\text{soft}}(\mathbf s_t, \mathbf a’)) d\mathbf a’,
\end{aligned}
\label{eq:soft_q_value}
\end{equation}
where $\mathcal{A}$ is the action space, and $\beta$ is a temperature parameter controlling the trade-off between entropy and reward. The optimal policy $\pi^*$ can be expressed in terms of the optimal Soft $Q$-function $Q_{\text{soft}}^*$ and optimal Soft $V$-function $V_{\text{soft}}^*$:
\begin{equation}
\small
\pi^*(\mathbf a_t \mid \mathbf s_t) = \exp\left(\frac{Q_{\text{soft}}^{*}(\mathbf s_t, \mathbf a_t) - V_{\text{soft}}^{*}(\mathbf s_t)}{\beta}\right).
\label{eq:optimal_relation}
\end{equation}

Building on this framework, \citet{rafailov2024r} formally proposed that an LM can be interpreted as an optimal soft $Q$-functions.
\begin{proposition}
    (Proof in Appendix ~\ref{app:proposition_proof}) In general maximum entropy reinforcement learning setting, a language model parameterized by $\pi_\theta$ can be seen as an optimal soft Q-function under some reward.
\end{proposition}
Thus, the corresponding optimal $Q$-function can be presented by $\pi_\theta$:
\begin{equation}
\small
    Q^\theta_\text{soft}(\mathbf s_t, \mathbf a_t) = \beta \log \pi_\theta(\mathbf a_t \mid \mathbf s_t) + V^\theta_\text{soft}(\mathbf s_t).
\label{eq:q_function_by_policy}
\end{equation}

\subsection{The DVO Objective}
Different from DPO, which optimizes language models at the response level using cross-entropy loss, we treat the policy as a $Q$-function and optimize it with Mean Squared Error (MSE) loss based on value estimations.

Specifically, following \citet{haarnoja2017reinforcement,schulman2017equivalence}, the $Q$-function can be optimized by using soft $Q$-learning:
\begin{equation}
\small
    J_Q(\theta) = \mathbb{E}_{t, \mathbf s_t, \mathbf a_t}\left[\frac{1}{2} \left(Q_{\text{soft}}^\theta(\mathbf s_t, \mathbf a_t) - y_t\right)^2\right],    
\end{equation}
where $y_t$ is the target $Q$-value. In one-step $Q$-learning, the target $Q$-value is computed as:
\begin{equation}
\small
y_t = R(\mathbf s_t,\mathbf a_t) + \hat{V}_{\text{soft}}^\theta(\mathbf s_{t+1}),
\label{eq:return_by_value}
\end{equation}
where $\hat{V}_{\text{soft}}^\theta(\mathbf s_{t+1})$ is the target $V$-function of $\mathbf s_{t+1}$. Substituting Eq.~\ref{eq:q_function_by_policy} and Eq.~\ref{eq:return_by_value} into the objective, we obtain:
\begin{equation}
\small
\begin{aligned}
J_Q(\theta)=\mathbb{E}_{t, \mathbf s_t, \mathbf a_t}\left[ \right. & \frac{1}{2}\left(\beta\log\pi_{\theta}(\mathbf{a}_t\mid \mathbf{s}_t)+V_{\text{soft}}^{\theta}(\mathbf{s}_t) \right.  \\
& \left. \left. -R(\mathbf s_t,\mathbf a_t) - \hat{V}_{\text{soft}}^\theta(\mathbf s_{t+1})\right)^2\right] \\
\end{aligned}
\end{equation}

Similar as DPO, the reward function $R(\mathbf s_t, \mathbf a_t)$ includes both the actual reward $r(\mathbf s_t, \mathbf a_t)$ and a KL penalty term:
\begin{equation}
\small
    R(\mathbf s_t, \mathbf a_t) = \beta \log \pi_\text{ref}(\mathbf a_t \mid \mathbf s_t) + r(\mathbf s_t, \mathbf a_t),
    \label{eq:reward_function}
\end{equation}
where $\pi_\text{ref}$ is a reference policy. To optimize the parameters $\theta$, we substitute the estimated value function $\hat{V}_{\text{soft}}^\theta(s_t)$ for $V_{\text{soft}}^\theta(s_t)$. This leads to the final objective:
\begin{equation}
\small
\begin{aligned}
    J_Q(\theta) &= \mathbb{E}_{t, \mathbf s_t, \mathbf a_t} [\frac{1}{2} \Bigg(\beta \log \frac{\pi_\theta(\mathbf a_{t} \mid \mathbf s_{t})}{\pi_{\text{ref}}(\mathbf a_{t} \mid \mathbf s_{t})} \\
    &\quad - \left( \hat{V}_{\text{soft}}^\theta(\mathbf s_{t+1})  +  r(\mathbf s_t, \mathbf a_t) - \hat{V}_{\text{soft}}^\theta(\mathbf s_{t})\right)\Bigg)^2]. \label{eq:final_objective}
\end{aligned}
\end{equation}

As a result, we can now directly align a language model with step-level value estimation, which is derived from outcome reward signals without requiring human labeling.

\textbf{Iterative Training.} This approach can be naturally extended into an iterative process. In each iteration, the latest policy model is used to generate new reasoning paths and corresponding value estimates, which are then used for further training.

\subsection{Target Value Estimation}
We investigate two representative methods for estimating target values for DVO training: MC estimation with a tree search algorithm~\citep{kocsis2006bandit,coulom2006efficient}, which directly approximates the value through simulated rollouts, and an outcome value model~\citep{yu2023outcome,feng2023alphazero}, which learns to predict the value based on the observed outcomes and states.

\subsubsection{Estimation Using MCTS}
We use Monte Carlo Tree Search (MCTS) to sample reasoning paths and estimate process value functions. Known for its success in AlphaZero and AlphaGo~\citep{silver2017mastering, silver2016mastering}, MCTS effectively balances exploration and exploitation while providing accurate state value estimates. It has also been widely adopted in complex reasoning tasks for data generation~\citep{xie2024monte,chen2024alphamath,feng2023alphazero,chen2024step}. In this framework, we modify MCTS to collect training data for the policy model $\pi_{\theta}$, incorporating the following key steps:

\textbf{Selection}: Starting from the root node, child nodes are recursively selected using the PUCT algorithm~\citep{rosin2011multi} until an expandable leaf node is reached. The next node is chosen based on a trade-off between exploration and exploitation:
\begin{equation*}
\small
    s' = \arg\max_{\mathbf a} [Q(\mathbf s,\mathbf a) + c \cdot \pi_\theta(\mathbf a \mid \mathbf s) \frac{\sqrt{N(\mathbf s)}}{1 + N(\mathbf s')}].
\end{equation*}

Here, $Q(\mathbf s,\mathbf a)$ represents the $Q$ value estimate from the tree search, $N(\mathbf s)$ is the visit count of node $\mathbf s$, and $c$ is a hyperparameter that regulates the trade-off between exploration and exploitation. 

\textbf{Expansion}: The selected node is expanded by adding the top-k most probable actions and their corresponding next states as child nodes.

\textbf{Evaluation \& Backup}: Leaf nodes are assigned a KL-regularized reward (Eq.~\ref{eq:reward_function}). During soft value function estimation, the entropy term from the policy model cancels out the KL regularization term, allowing the use of the original reward for sampling. Terminal node results are propagated back to the root node, updating visit counts and cumulative rewards:
\begin{equation*}
\small
\begin{aligned}
    & N(s', a) \leftarrow N(s, a) + 1 \\
    & V(s) \leftarrow \frac{\sum_a N(s')(r(s, a)+V(s'))}{\sum_a N(s')}
\end{aligned}
\label{eq:reward_model_equation}
\end{equation*}

\subsubsection{Estimation Using a Value Model}
Outcome value models are widely used in inference-time tree search, providing value guidance for each step in the search process~\cite{yu2023outcome,wang2023math}. We leverage a value model to assist in training the policy model $\pi_\theta$. During value model training, we minimize a mean squared error loss to directly model the outcome reward:
\begin{equation*}
\small
    L_{V}(\phi) = \mathbb{E}_{(s_i^{(n)}, r_i)\in D} \left[\frac{1}{n} \sum_{t=1}^{n} \left\| V_\phi(\mathbf{s}_t) - r_i \right\|_2^2\right],
\end{equation*}
where $s_i^{(n)}$ represents a trajectory of $n$ states, $r_i$ is the associated outcome reward, and $V\phi(\mathbf{s}_t)$ is the predicted value for $\mathbf{s}_t$ by the value model parameterized by $\phi$. The data is defined as $D = \{(s_i^{(n)}, r_i)\}$, which is responses generated from $\pi_\theta$. Once trained with $D$, the value model provides target values to update the policy model $\pi_\theta$. Notably, it remains fixed during policy optimization and is not updated alongside $\pi_\theta$.

\section{Experimental Settings}

\noindent \textbf{Datasets.}
We validate our approach on both math word problems and commonsense reasoning tasks. For math word problems, we use GSM8K~\citep{cobbe2021training} and MATH~\citep{hendrycks2021measuring} as in-domain test sets. The former contains over 1,300 grade school math problems, while the latter includes 5,000 problems spanning five difficulty levels. AGIEval-Math~\citep{zhong2023agieval} is used as an out-of-distribution (OOD) test set. For training, we take MetaMath~\citep{yu2023metamath}, a dataset with 395K math problems, solutions, and correct answers. In our training process, only the correct answers are used, and solutions are excluded. During training, we sample 10,000 problems per round.

\noindent \textbf{Models.}
We employ three models as our backbones: two state-of-the-art general language model in different size, Llama3-8B-Instruct, and Llama3-70B-Instruct~\citep{dubey2024llama}, and a specialized math-focused language model, DeepSeek-Math-Instruct-7B~\citep{shao2024deepseekmath}.

\noindent \textbf{Implementation Details.}
Before DVO training, we fine-tune the model backbones to ensure proper formatting of reasoning steps, with ``\verb|\|n\verb|\|n'' used as a delimiter between steps and ``Final Answer:'' marking the conclusion. Our approach focuses on using MCTS for target value collection. All models are optimized over three rounds of DVO training. In each round, we perform a modified MCTS for each problem, where each node represents a reasoning step generated by the model, separated by ``\verb|\|n\verb|\|n'' The tree search is configured with a maximum of 80 iterations, a breadth limit of 5, and a depth limit of 8. From the search tree, we select four positive and four negative examples for policy updates. During evaluation, all assessments are conducted in a zero-shot setting. Details of the fine-tuning process and evaluation setting can be found in App.~\ref{app:implementation_details}.

\begin{figure}[t]
    \centering\small
    \setlength{\abovecaptionskip}{6pt}  
    \includegraphics[width=0.6\linewidth]{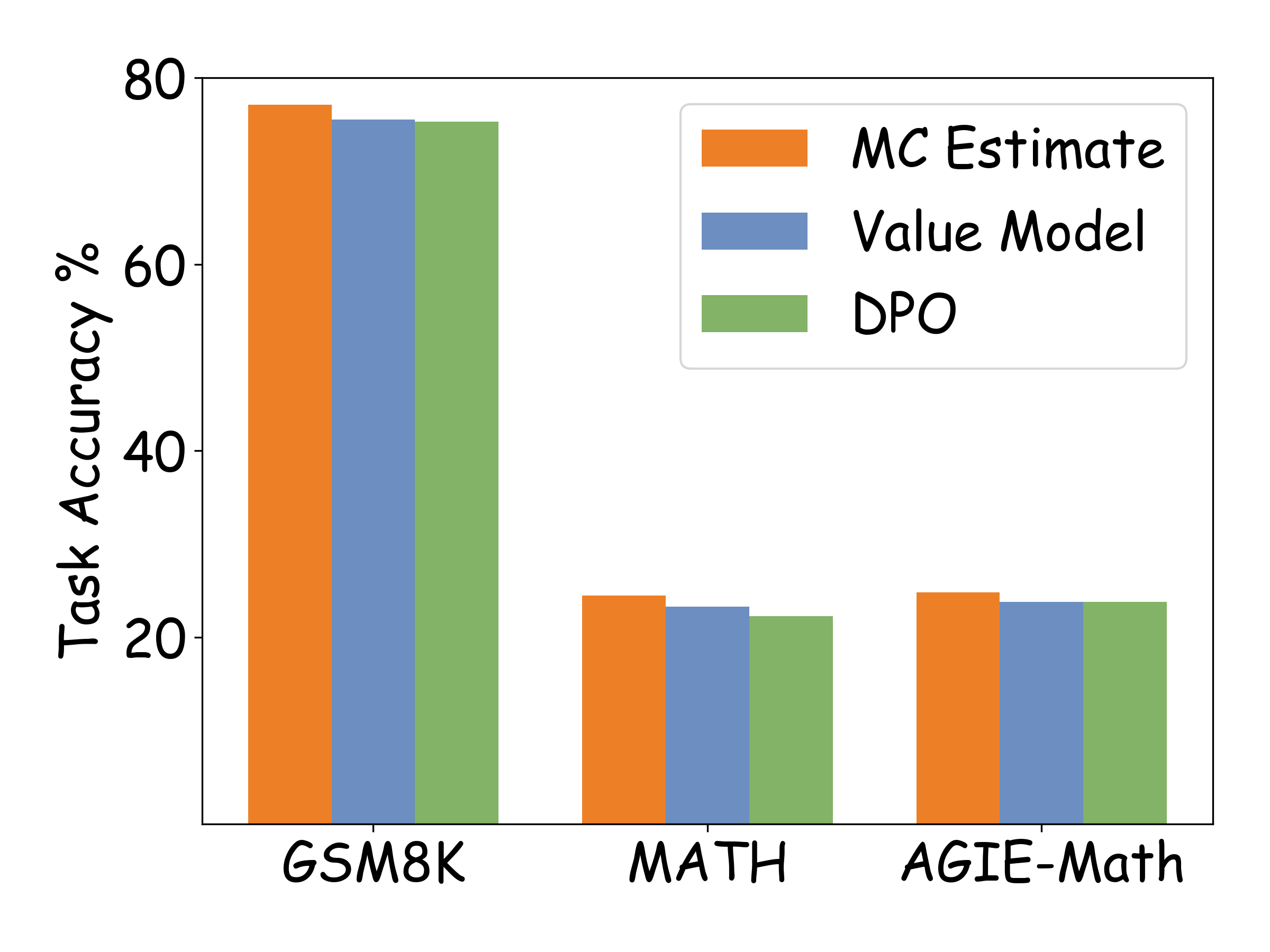}
    \caption{Different value estimation in DVO.}
    \label{fig:value_estimate_comparison}
    \vspace{-5pt}
\end{figure}

\noindent \textbf{Baselines.}
We compare our methods with several baselines which leverage self-generated data to boost the reasoning performance of LLMs, which are categorized based on how they train the policy model: (i) DPO~\citep{rafailov2023directpreferenceoptimizationlanguage,yuan2024advancing} samples positive and negative paths from the results of MCTS to construct pairs and use DPO to train the policy model at the solution level. (ii) Step-DPO~\citep{zhang2024chain,lai2024step,setlur2024rl} is a step-level DPO algorithm, aligning preferences at a finer granularity by constructing pair comparisons at the step level and training the policy model using DPO. (iii) KTO~\citep{ethayarajh2024kto} is an extension of DPO, which trains the policy model without relying on paired data. (iv) RFT~\citep{chen2024alphamath,feng2023alphazero,yuan2023scaling} directly samples correct paths from the tree and use supervised fine-tuning to train the policy model.   In our comparative experiments, we use the same trees in the first round to ensure consistency across methods. Further details are provided in App.~\ref{app:comparative_exp_details}.

\section{Results}
\subsection{Target Value Estimation}

We first compare the two options for target value estimation: MC estimation and outcome value model. Specifically, we initialize the value network with Llama-3-8B-Instruct, following the OVM approach~\citep{yu2023outcome} to train the network, and use it to label target values for DVO training data. The results, shown in Fig.~\ref{fig:value_estimate_comparison}, indicate that the MC estimation achieves better performance while avoiding the additional computational overhead required to train a separate value network. 

Therefore, all subsequent experiments are performed using MCTS for value estimation.

\subsection{Main Results}
\begin{table}[t]
\centering\scriptsize
\begin{tabular}{@{}l@{\hspace{2pt}}|l|cc@{\hspace{5pt}}|c@{}}
\toprule
\multicolumn{1}{l|}{\multirow{2}{*}{\bf Model}} & \multirow{2}{*}{\bf Size} & \multicolumn{2}{c|}{\bf In-domain}                        & \bf Out-of-domain \\
\cmidrule(l){3-5} 
&                       & \bf GSM8K & \bf MATH & \bf AGIE-Math\\
\midrule
DeepSeek-Math-7B-Ins.$^\dagger$            & 7B                    & 81.4                      & 44.4                      & 43.4                              \\
+ RFT                    & 7B                    & 81.0                      & 44.1                      & 43.5                               \\
+ DPO               & 7B                    & 82.1                      & \underline{46.0}                      & \underline{44.8}                               \\
+ KTO             & 7B                    & \underline{83.0}                      & 44.0                      & 43.2                               \\
+ Step-DPO             & 7B                    & 82.0                      & 45.7                      & 43.3                               \\
+ DVO (Ours)                            & 7B                    & \bf 85.2                      & \bf 47.6                      & \bf 46.3\\
\midrule
Llama-3-8B-Instruct$^\dagger$                    & 8B                    & 74.6                      & 22.5                      & 23.5                               \\
+ RFT                    & 8B                    & 75.6                      & 23.4                      & 23.9                               \\
+ DPO               & 8B                    & 75.3                      & 22.3                      & 23.8                               \\
+ KTO             & 8B                    & 76.0                      & \underline{24.1}                      & \underline{24.3}                               \\
+ Step-DPO             & 8B                    & \underline{76.6}                      & 23.3                      & 23.5                               \\
+ DVO (Ours)                            & 8B                    & \bf 80.6                      & \bf 26.5                      & \bf 27.9                               \\
\midrule
Llama-3-70B-Instruct$^\dagger$                   & 70B                   & 85.4                      & 34.9                      & 33.8                               \\
+ RFT                    & 70B                   & 86.3                      & 34.8                      & 35.2                               \\
+ DPO               & 70B                   & 82.7                      & 32.6                      & 31.5                               \\
+ KTO             & 70B                   & 86.5                      & 35.8                      & \underline{35.5}                               \\
+ Step-DPO             & 70B                   & \underline{87.7}                      & \underline{36.7}                      & 35.0                               \\
+ DVO (Ours)                            & 70B                   & \bf 90.7                      & \bf 40.3                      & \bf 38.9                               \\
\bottomrule
\end{tabular}
\caption{Results on math reasoning tasks. $^\dagger$~indicates that the model has been fine-tuned to follow step-by-step pattern. Highest performance results in comparison are highlighted in \textbf{bold}, and the second-high results are \underline{underlined} for clarity.}
\label{tab:main_DVO_math.result}
\vspace{-5pt}
\end{table}
\begin{table}[t]
\centering\scriptsize
\begin{tabular}{@{}l|ccc|c@{}}
\toprule
\multicolumn{1}{l|}{\multirow{2}{*}{\bf Model}} & \multicolumn{3}{c|}{\bf In-domain}                        & \multicolumn{1}{c}{\bf Out-of-domain} \\
\cmidrule(l){2-5} 
\multicolumn{1}{l|}{}                           & \multicolumn{1}{c}{\bf OBQA} & \multicolumn{1}{c}{\bf CSQA} & \multicolumn{1}{c|}{\bf ARC-C} & \multicolumn{1}{c}{\bf SciQ}\\
\midrule
Llama-3-8B-Instruct$^\dagger$                    & 81.2                     & 74.1                     & 79.7                       & 90.7                              \\
+ RFT                    & 79.6                     & 77.0                     & 73.9                       & 87.0                              \\
+ DPO               & 82.8                     & 78.7                     & 80.9                       & 90.6                              \\
+ KTO             & 83.4                     & 79.0                     & 81.4                       & 91.5                              \\
+ Step-DPO             & 82.2                     & 79.0                     & \bf 82.0                       & 91.6                              \\
+ DVO (Ours)                            & \bf 84.2                     & \bf 79.3                     & 81.7                       & \bf 92.4                              \\
\bottomrule
\end{tabular}
\caption{Results on commonsense reasoning tasks.  Highest results are highlighted in \textbf{bold}. }
\label{tab:main_DVO_general.result}
\vspace{-5pt}
\end{table}

We compare DVO with baseline methods on math reasoning tasks. As shown in Tab.~\ref{tab:main_DVO_math.result}, DVO consistently delivers stable gains across various model backbones and significantly outperforms DPO and its alternatives KTO and Step-DPO. Specifically, DVO improves DeepSeek-Math from 83.0\% (KTO) to 85.2\%, Llama-3-8B from 76.6\% (Step-DPO) to 80.6\%, and Llama-3-70B from 87.7\% (Step-DPO) to 90.7\% on GSM8K. On the more challenging MATH benchmark, these models achieve improvements of 1.6\%, 2.7\%, and 4.1\% on the models, respectively. Notably, the performance gains are observed across different model sizes, highlighting the scalability of DVO. Moreover, all improvements are achieved without the use of external annotations, relying solely on final answer evaluations, demonstrating the efficiency and simplicity of the DVO framework.

On two out-of-domain (OOD) datasets, DVO achieved notable improvements. For Llama-3-70B-Instruct, the accuracy on AGI-Eval-Math improved from 33.8\% to 38.9\%. Similar gains were observed on other backbones, showcasing its robust generalization capability.

\subsection{Commonsense Tasks}
We further validate the effectiveness of DVO on commonsense reasoning tasks. The results are presented in Tab.~\ref{tab:main_DVO_general.result}. The experiments are conducted on Llama-3-8B-Instruct (see implementation details in App.~\ref{app:commonsense_reasoning_exp}). DVO demonstrates strong performance and generalization capabilities in commonsense reasoning tasks, further highlighting its versatility across different reasoning paradigms.

\subsection{Ablation Study}
\begin{figure}[t]
    \centering
    \setlength{\abovecaptionskip}{6pt}
    \begin{subfigure}[t]{0.5\linewidth}
        \centering
        \includegraphics[width=\linewidth]{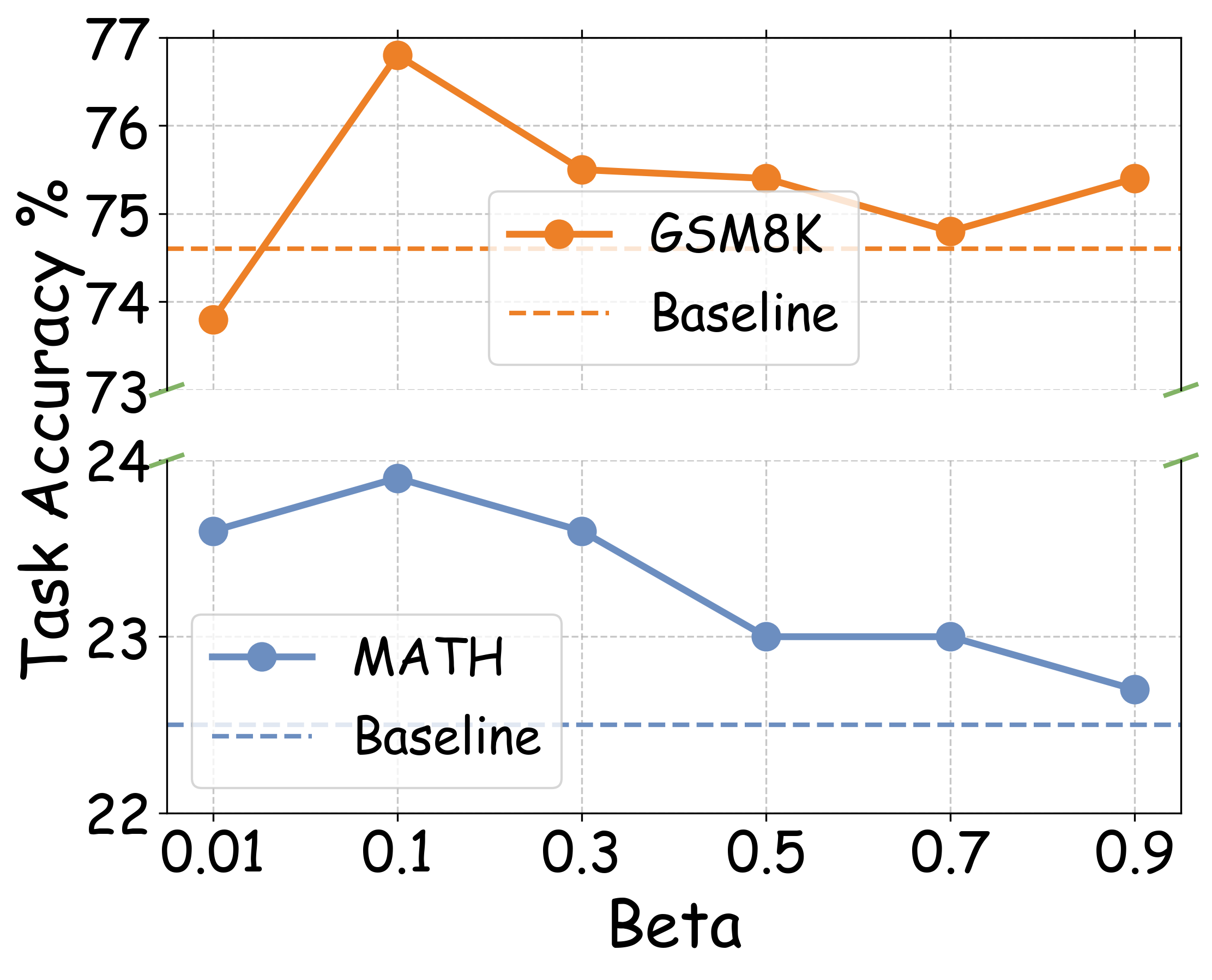}
        \caption{}
        \label{fig:beta_scan}
    \end{subfigure}%
    \hfill
    \begin{subfigure}[t]{0.5\linewidth}
        \centering
        \includegraphics[width=\linewidth]{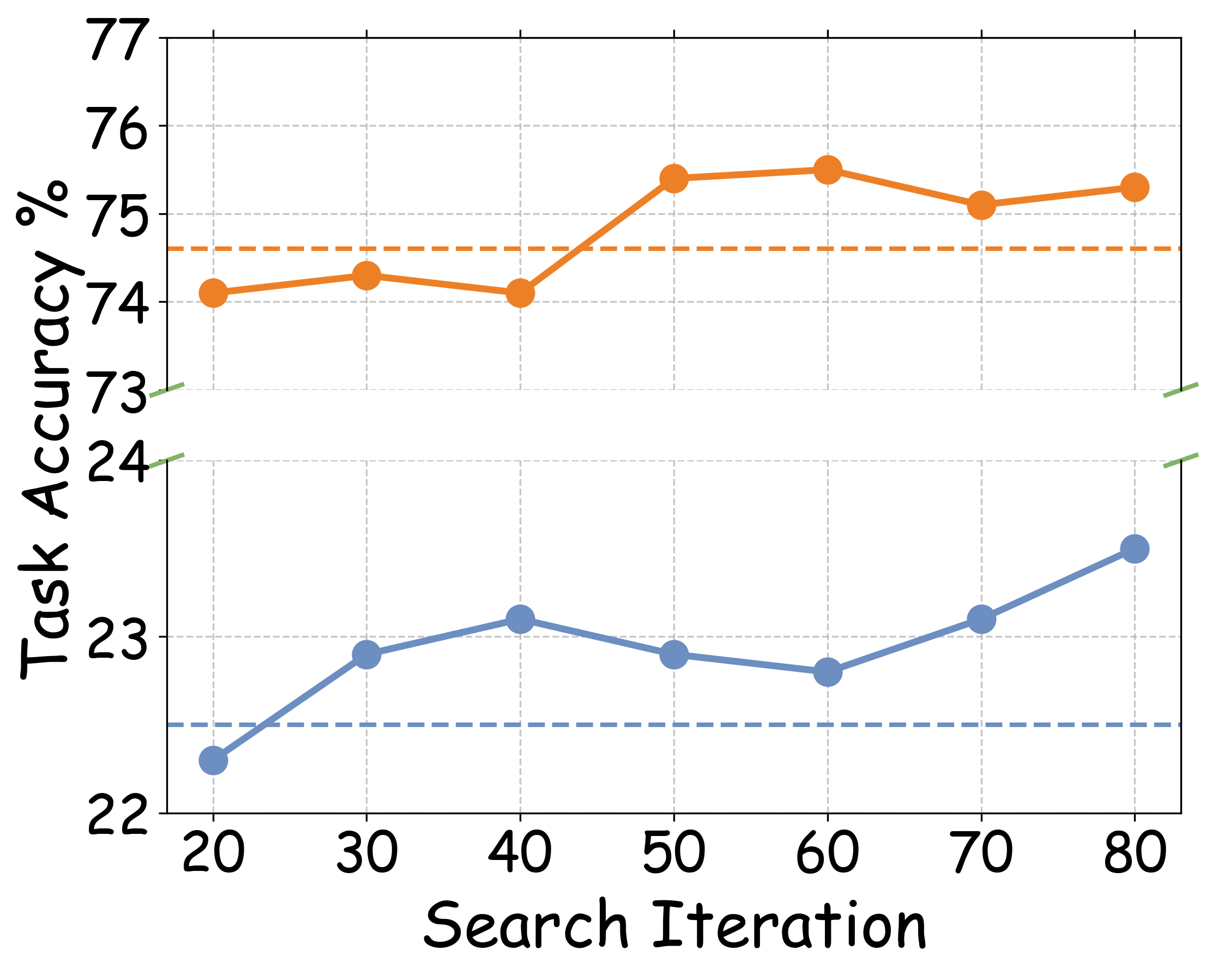}
        \caption{}
        \label{fig:iteration_scan}
    \end{subfigure}
    \caption{Ablation study on hyperparameters. The left figure demonstrates the effect of varying $\beta$ values during training, while the right figure highlights the impact of search iterations in MCTS.}
    \label{fig:beta_and_iteration}
\end{figure}
\begin{figure}[t]
    \centering
    \setlength{\abovecaptionskip}{6pt}
    \begin{subfigure}[t]{0.5\linewidth}
        \centering
        \includegraphics[width=\textwidth]{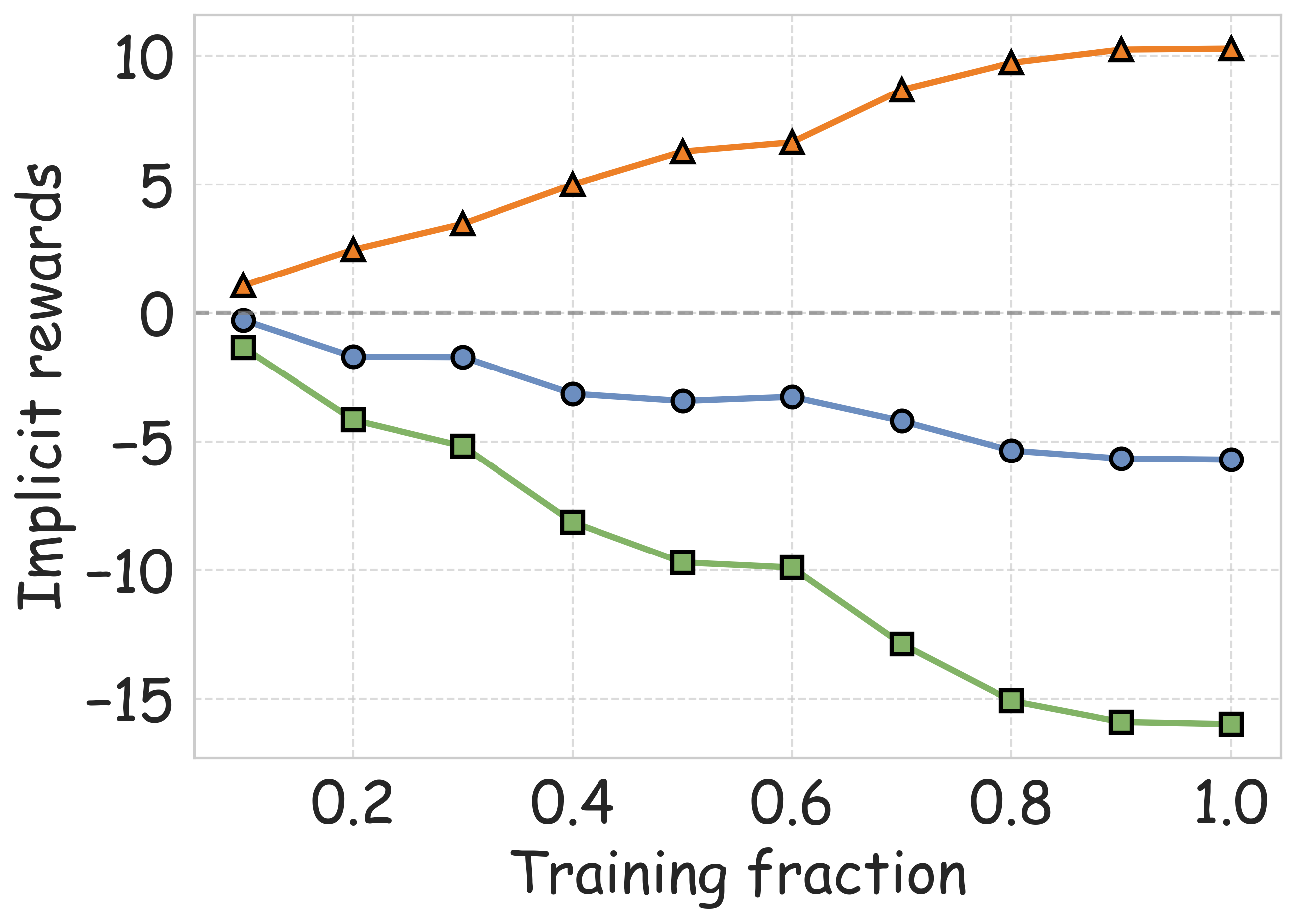}
        \caption{DPO.}
        \label{fig:reward_evolution_a}
    \end{subfigure}%
    \hfill
    \begin{subfigure}[t]{0.5\linewidth}
        \centering
        \includegraphics[width=\textwidth]{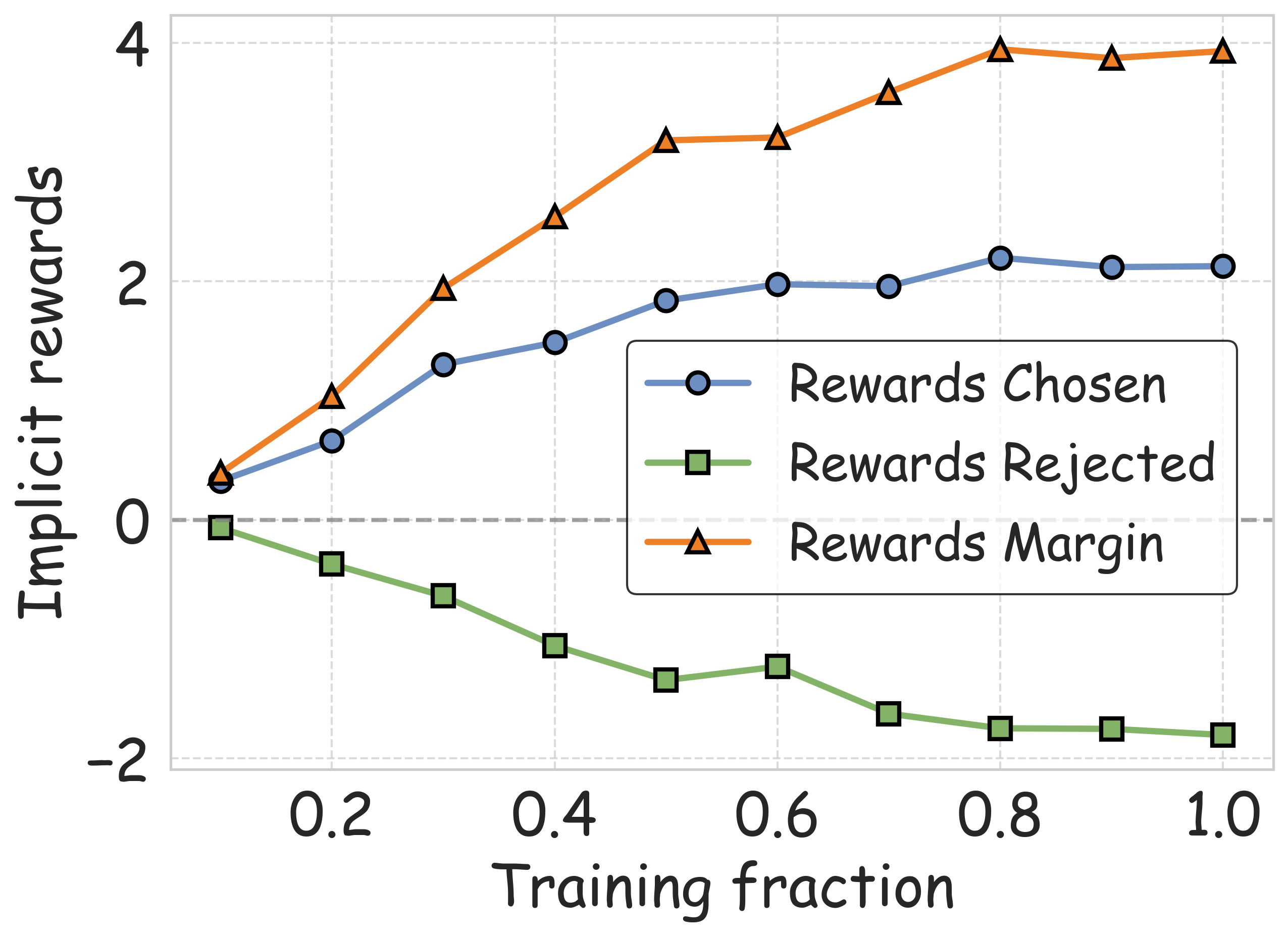}
        \caption{DVO.}
        \label{fig:reward_evolution_b}
    \end{subfigure}
    \caption{volution of implicit rewards during DPO and DVO training. While the reward margin increases, the implicit reward of positive solutions decreases in DPO but increases in DVO.}
    \label{fig:implicit_reward_revolution}
    \vspace{-5pt}
\end{figure}

\paragraph{Ablation on beta $\beta$.} 
We evaluate the impact of $\beta$ on DVO performance by scanning different values on the same training set, with all models trained for a single epoch. As shown in Fig.~\ref{fig:beta_scan}, DVO achieves the best performance when $\beta=0.1$. Both excessively large and small values of $\beta$ result in degraded performance, indicating that DVO is sensitive to the balance between KL divergence regularization and entropy constraints.

\paragraph{Ablation on Search Iteration.}
We evaluate the impact of the number of search iterations in MCTS on performance. As the number of iterations increases, the accuracy of value estimation improves. As shown in Fig.~\ref{fig:iteration_scan}, inaccurate value estimation leads to a noticeable drop in performance. Additionally, we observe that on the simpler GSM8K dataset, the model requires higher precision in value estimation. We hypothesize that this is because simpler tasks demand finer distinctions between different steps to improve reasoning performance. Overall, we find that search iterations beyond 40 generally lead to better results.

\begin{figure*}[t]
    \centering\small
    \setlength{\abovecaptionskip}{6pt}  
    \includegraphics[width=1.0\textwidth]{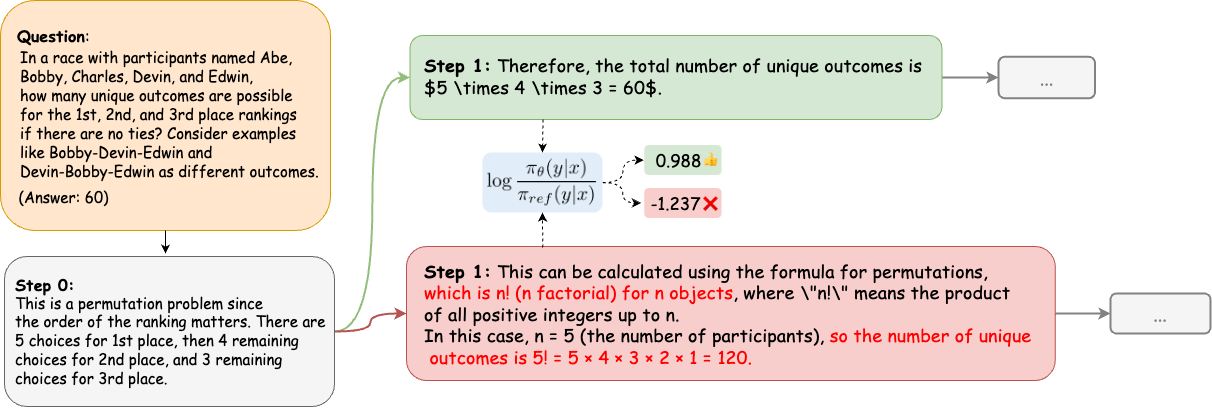}
    \caption{Step-by-step analysis of solutions generated by the reference model. The figure compares correct (top) and incorrect (bottom) CoT responses sampled from $\pi_{\text{ref}}$. Each action is evaluated based on the DVO credit defined before, with DVO credit assigned positively for correct steps and negatively for incorrect steps.}
    \vspace{-5pt}
    \label{fig:value_estimate_case_study}
\end{figure*}
\begin{figure}[t]
    \centering\small
    \setlength{\abovecaptionskip}{6pt}  
    \includegraphics[width=0.6\linewidth]{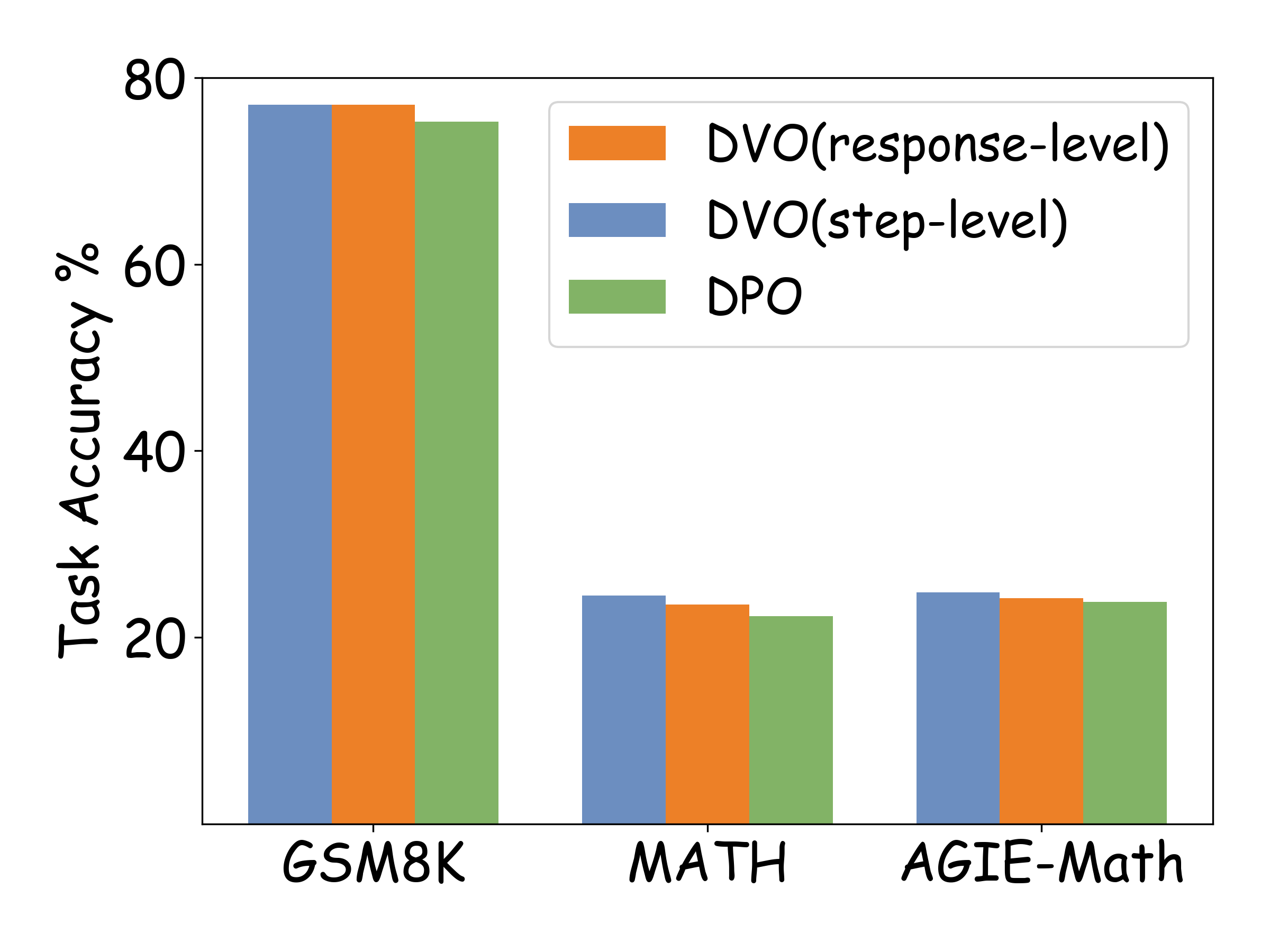}
    \vspace{-5pt}
    \caption{Different supervision granularity in DVO.}
    \label{fig:value_granularity_comparison}
\end{figure}

\subsection{Analysis}
\paragraph{Deep observation on implicit reward.}
Recent work~\citep{chen2024noise,yuan2024advancing,pal2024smaug} has shown that during DPO training, while the implicit reward margin between positive and negative responses continues to increase, the implicit reward of the positive responses gradually decreases (as shown in Fig.~\ref{fig:reward_evolution_a}). 
This phenomenon likely arises because DPO’s optimization objective emphasizes maximizing the reward gap between positive and negative responses, rather than directly increasing the rewards for positive responses or decreasing those for negative ones. Consequently, DPO causes a decrease in the probability of positive data, resulting in suboptimal outcomes.
We analyze the changes in implicit rewards during DVO training, as shown in Fig.~\ref{fig:reward_evolution_b}. Unlike DPO, we observe that the implicit rewards for the chosen responses increase steadily throughout training. This behavior is likely attributed to DVO’s objective of directly aligning value functions, which effectively enhances performance on reasoning tasks by maintaining a consistent focus on improving the rewards for preferred responses.

\paragraph{Step level vs. response level DVO.}

DVO can be applied at the response level, where the entire response is treated as a single action, and optimization is performed over the entire response. This setup is similar to DPO, while the key difference is that DVO directly aligns the $Q$-value of the response rather than employing contrastive learning on paired data. We conduct comparative experiments on Llama-3-8B-Instruct, and the results are shown in Fig.~\ref{fig:value_granularity_comparison}. As the figure shows, response-level optimization underperforms step-level optimization, demonstrating the effectiveness of finer-grained supervision for improving reasoning. Notably, response-level optimization still outperforms DPO, suggesting that at the same granularity, optimizing for explicit value signals rather than preference relationships enables the policy model to generate high-quality responses more effectively.

\textbf{Credit assignment cases.}
Similar as ~\citet{rafailov2024r}, after DVO training we can take $\frac{\pi_\theta(a_t \mid s_t)}{\pi_{\text{ref}}(a_t \mid s_t)} = r(s_t, a_t) + V_\theta(s_{t+1}) - V_\theta(s_t)$ from Eq.~\ref{eq:final_objective} as the DVO credit, indicating the potential reward or loss from transitioning between states. We sample correct and incorrect reasoning chain from the reference model and visualize steps with the corresponding DVO credit. As illustrated in Fig.~\ref{fig:value_estimate_case_study}, each step is colored based on the DVO credit, with erroneous steps receiving low credit (darker colors). This example demonstrates that DVO has learned step-level value information, enabling it to identify faulty reasoning steps.

\section{Conclusion}
We introduced Direct Value Optimization (DVO), a novel framework that enhances LLMs’ reasoning ability by directly optimizing reasoning paths with estimated step-wise value estimation instead of preference labels. This approach provides finer-grained supervision signals while eliminating the need for paired data construction. Practically, we generate offline samples via tree search and estimate target values using either MC estimation or a value model, then update the policy model with MSE loss.
Extensive experiments on both in-domain and out-of-domain datasets demonstrate DVO’s effectiveness. Notably, these improvements were achieved without requiring new query data or additional supervision. Furthermore, our analytical experiments validated the utility of step-level value information, underscoring the broader applicability of DVO across various reasoning tasks.

\newpage
\section*{Limitations}
The proposed DVO approach exhibits two primary limitations.
First, the current implementation of DVO is confined to offline settings. However, this limitation can be addressed in future work by integrating the MCTS search process with the update step in the training pipeline, thereby enabling online operation.
Second, due to computational constraints, we restrict the MCTS search width to 5 in our experiments.
Theoretically, with adequate computational resources, expanding the MCTS search width would yield superior performance through two mechanisms: we will get a more accurate value estimate, while simultaneously the scale of data available for training is also larger.

\bibliography{custom}

\appendix

\newpage
\setcounter{proposition}{0}
\section{Proof of Proposition}
\label{app:proposition_proof}
\begin{proposition}
    In general maximum entropy reinforcement learning setting, a language model parameterized by $\pi_\theta$ can be seen as an optimal soft Q-function under some reward.
\end{proposition}

\begin{equation}
\small
    \pi_\theta(\mathbf a_t \mid \mathbf s_t) = \exp\left(\frac{1}{\beta} \left(Q^*(\mathbf s_t, \mathbf a_t) - V^*(\mathbf s_t)\right)\right),
\end{equation}

\begin{proof}
The proposition is introduced in \citet{rafailov2024r}, and we provide a similar proof here at the step level. Firstly, we prove that a large language model in a token-level MDP represents an optimal soft $Q$-function. In the token-level MDP, each action $\mathbf a_t$ corresponds to a token, and the action space $\mathcal{A}$ is defined as the vocabulary of the language model. At step $t$, the state $\mathbf s_t$ comprises a prompt and a sequence of generated reasoning tokens. The policy model $\pi_\theta$ selects the next action $\mathbf a_t$ based on the probability distribution $\pi_\theta(\mathbf a_t \mid \mathbf s_t)$. To encourage robust reasoning, the reward function $r(\mathbf s_t, \mathbf a_t)$ is defined at the token level. The transition function $f$ concatenates the current state $s_t$ with the selected action $\mathbf a_t$, producing the next state $f(\mathbf s_t, \mathbf a_t) = \mathbf s_t \mid \mathbf a_t$. The discount factor $\gamma$ is set to 1. We also omit explicit references in transition probabilities $P(\mathbf s_{t+1} \mid \mathbf s_t, \mathbf a_t)$

A language model can be represented as a set of logits $\ell_\theta(\mathbf s)$, which, combined with a temperature parameter $\beta$, define the action distribution via a softmax function:
\begin{equation}
\small
\pi_\theta(\mathbf a \mid \mathbf s) = \frac{\exp(\ell_\theta(\mathbf s)[\mathbf a]/\beta)}{\sum_{\mathbf a’ \in \mathcal{A}} \exp(\ell_\theta(\mathbf s)[\mathbf a’]/\beta)}.
\label{eq:lm_policy}
\end{equation}
The optimal $Q$-function $Q^*(\mathbf s_t, \mathbf a_t)$ is defined as the expected cumulative reward when starting from state $\mathbf s_t$ and taking action $\mathbf a_t$, following the Bellman equation:
\begin{equation}
\small
\begin{aligned}
    Q^*(\mathbf s, \mathbf a) &= r(\mathbf s, \mathbf a) + \beta \log V^*(\mathbf s^\prime)\\
    &= r(\mathbf s, \mathbf a) + \beta \log \sum_{\mathbf a’ \in \mathcal{A}} \exp\left(\frac{1}{\beta} Q^*(\mathbf s’, \mathbf a’)\right)
\end{aligned}
\label{eq:bellman_optimal_q}
\end{equation}
where $s^{\prime} = f(\mathbf s, \mathbf a)$ is the next state. The optimal policy $\pi^*(\mathbf a \mid \mathbf s)$ is derived from the optimal $Q$-function:
\begin{equation}
\small
\pi^*(\mathbf a \mid \mathbf s) = \frac{\exp(Q^*(\mathbf s, \mathbf a)/\beta)}{\sum_{\mathbf a’ \in \mathcal{A}} \exp(Q^*(\mathbf s, \mathbf a’)/\beta)}.
\label{eq:optimal_policy}
\end{equation}
Given the LM policy in Eq.~\eqref{eq:lm_policy}, let $Q^*(\mathbf s, \mathbf a) \triangleq \ell_\theta(\mathbf s)[\mathbf a]$. Substituting $Q^*(\mathbf s, \mathbf a)$ into Eq.\eqref{eq:optimal_policy}, we obtain:
\begin{equation}
\small
\pi_\theta(\mathbf a \mid \mathbf s) = \frac{\exp(Q^*(\mathbf s, \mathbf a)/\beta)}{\sum_{\mathbf a’ \in \mathcal{A}} \exp(Q^*(\mathbf s, \mathbf a’)/\beta)}.
\end{equation}
Thus, the LM policy $\pi_\theta(\mathbf a \mid \mathbf s)$ matches the optimal policy $\pi^*$ derived from $Q^*(\mathbf  s, \mathbf a)$ in Eq.~\eqref{eq:optimal_policy}. To ensure consistency with the Bellman equation, the reward function $r(\mathbf s, \mathbf a)$ must satisfy:
\begin{equation}
\small
r(\mathbf s, \mathbf a) = \ell_\theta(\mathbf s)[\mathbf a] - \beta \log \sum_{\mathbf a’ \in \mathcal{A}} \exp\left(\frac{1}{\beta}\ell_\theta(\mathbf s’)[\mathbf a’]\right),
\label{eq:reward_consistency}
\end{equation}
where $\mathbf s^{\prime} = f(\mathbf s, \mathbf a)$. This ensures that the logits $\ell_\theta(\mathbf s)$ represent the optimal $Q$-function. Thus the logits of the language model correspond to the optimal $Q$-function, and the resulting policy is optimal for this setting. 

While the above argument is presented at the token level, an equivalent view is to treat multiple tokens as a single “macro-action” and define the step-level reward as the sum of the token-level rewards for those tokens. In this step-level perspective, one action $\widetilde{\mathbf a}=(\mathbf a_1,\mathbf a_2,\dots,\mathbf a_L)$ corresponds to sequentially choosing $\mathbf a_1,\dots,\mathbf a_L$ at the token level. If we denote by $\pi_\theta$ the token-level optimal policy, then the probability of executing the macro-action $\widetilde{\mathbf a}$ under $\pi_\theta$ is precisely the product of its single-token probabilities: $\pi_\theta(\mathbf a_1 \mid \mathbf{s}) \times \cdots \times \pi_\theta(\mathbf a_L \mid \mathbf{s}^{(L-1)})$. Consequently, viewing several tokens as one step does not alter the overall sequence distribution or the value derived from it, the language model’s logits still induce the same optimal strategy under this coarser step-level view, since the probability of any generated block of tokens and its accumulated reward remain consistent with the original token-level formulation.
\end{proof}

\section{Implementation Details}
\label{app:implementation_details}
To enable LLMs to follow Step-by-step pattern, we collect self-generated data to fine-tune all backbone models. Specifically, similar to~\citet{lightman2023letsverifystepstep}, we start by generating multiple solutions to 10,000 MetaMath problems in a few-shot manner and use the correct solutions to SFT the backbone models. The objective here is to avoid introducing external reasoning paths, as our focus is solely on leveraging answers to enhance the model’s performance. The few-shot prompt is shown as below:
\noindent
\begin{tabular*}{\linewidth}{l} 
    \small
    \textbf{Few-shot Prompt} \\
    \hline
    You are tasked with solving the provided math \\ word problem by following these instructions: \\ 1. Formulate and solve the equations using \\ algebraic methods, ensuring each equation is \\ written strictly in LaTeX syntax. \\ 2. Document each step of your process clearly. \\ Use double line breaks '\verb|\n\n|' to separate each \\ step and ensure that there are no double line \\ breaks within a single step. \\ 3. Ensure the number of reasoning steps is within \\ 8 steps. \\ 4. Conclude your solution by stating the final \\ answer after the phrase 'Final Answer:'. \\  \\ \textless\textbar Begin Question \textbar\textgreater \\ \{\{Question1\}\} \\ \textless\textbar End Question \textbar\textgreater \\  \\ \textless\textbar Begin Answer\textbar\textgreater \\ \{\{Answer1\}\} \\ \textless\textbar End Answer \textbar\textgreater \\ \\ ... $\times$ n shots \\  \\ \textless\textbar Begin Question \textbar\textgreater \\ \{\{QUESTION\}\} \\ \textless\textbar End Question \textbar\textgreater \\  \\ \textless\textbar Begin Answer\textbar\textgreater
    \\\hline
\end{tabular*}
\vspace{3mm}

For each problem, we generate 64 completions and filter out responses with correct answers to train the backbone models. We then fine-tune all backbones on the self-generated data for one epoch, using a temperature of $2e{-5}$ and a batch size of 128.

For evaluation, we set the temperature to $0.7$ and run three test trials in zero-shot setting, reporting the average result.

\section{Comparative Experiments Implementation Details}
\label{app:comparative_exp_details}
To facilitate a comprehensive comparison of the DVO with other baseline algorithms, we implement all algorithms on the same dataset. The training data are sampled from the data collected in the first round. The sampling methods vary according to the specific algorithm, detailed as follows:
\begin{itemize}
    \item RFT: Samples positive instances from the trees based on the solution correctness, with a maximum of four positive instances per question.
    \item DPO: Samples paired solutions based on the solution correctness, with each question allowing for a maximum of four pairs.
    \item KTO: Samples both positive and negative instances based on the solution correctness, allowing for up to four positive solutions and four negative solutions per question.
    \item Step-DPO: Conducts step-level paired sampling of positive and negative instances based on the step value estimation, with a maximum of four pairs per question.
    \item DVO: Similar to KTO, it samples both positive and negative instances based on the step value estimation, with up to four positive solutions and four negative solutions per question.
\end{itemize}

All models are trained with the same set of hyperparameters. $\beta$ is fixed at $0.1$, $\lambda_+/\lambda_-$ is set at $1.33$ for KTO, and a learning rate of $5 \times e^{-7}$ with a cosine scheduler and 0.1 warmup ratio. Training is conducted over 3 epochs.

\label{app:commonsense_reasoning_exp}
For commonsense reasoning, we evaluate on OBQA~\cite{OpenBookQA2018}, CSQA~\cite{talmor2022commonsenseqa}, and ARC-Challenge~\cite{clark2018think} as in-domain test sets, with SciQ~\cite{welbl2017crowdsourcing} serving as the OOD test set. The training data is a combination of the training sets from three datasets, with approximately 7,000 examples used per round. Similar to the math reasoning tasks, we first fine-tune the backbone model on commonsense reasoning data, followed by DVO training. The data construction and training hyperparameters are identical to those used in the math reasoning tasks, with the only difference being that we performed a single round of training to validate the effectiveness of DVO.

\newpage
\begin{figure}[!htbp]
    \centering\small
    \setlength{\abovecaptionskip}{6pt}  
    \includegraphics[width=0.8\linewidth]{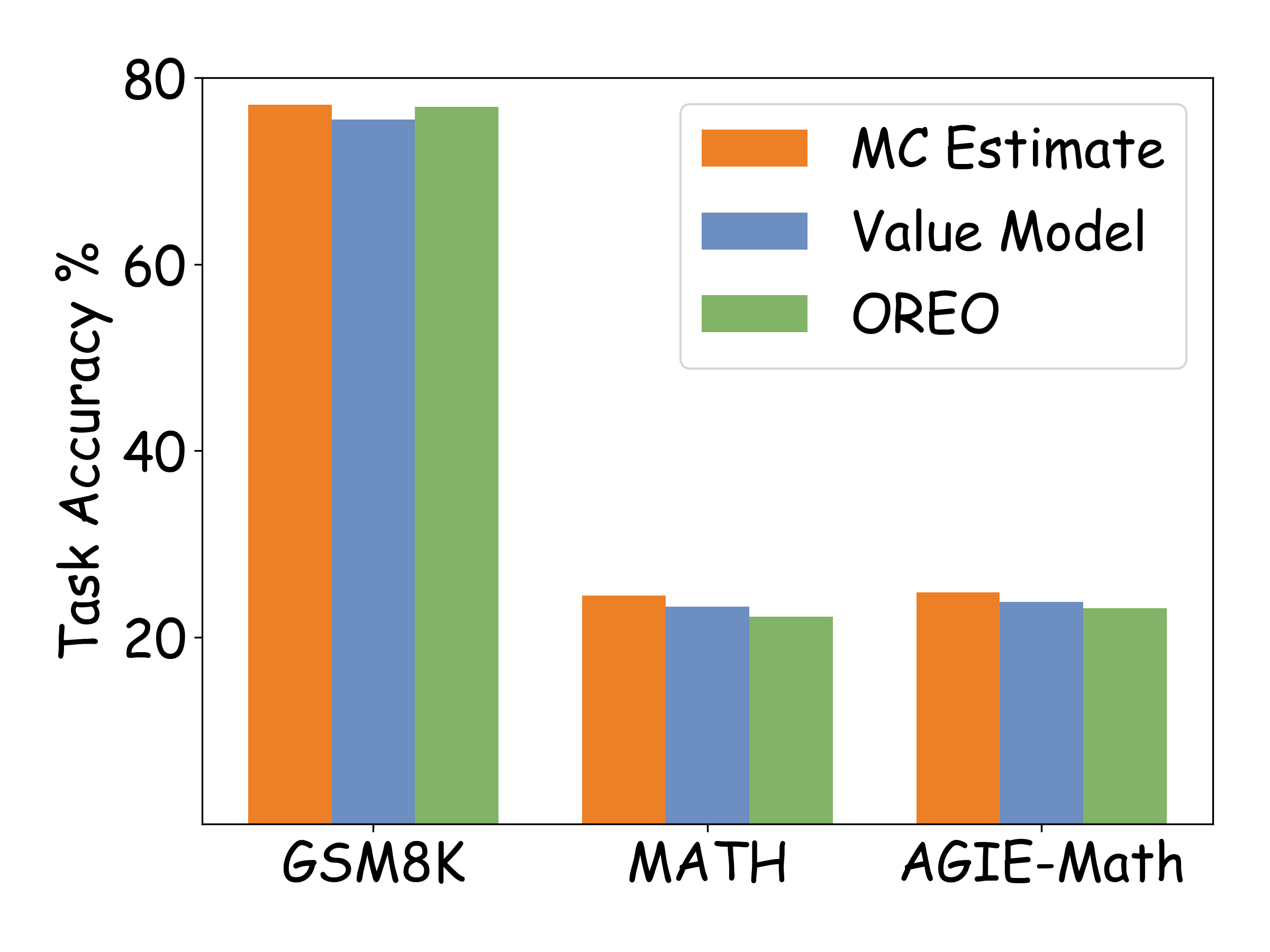}
    \caption{Compare DVO with OREO.}
    \label{fig:value_estimate_comparison_oreo}
\end{figure}

\section{Compare with OREO}
\label{app:compare_with_oreo}
Similar to DVO, the recent work OREO~\cite{wang2024offline} extends its objective from maximum entropy reinforcement learning. However, unlike DVO, OREO requires joint training of a separate value model and introduces a KL regularization term. We present the results on the mathematical reasoning task in Fig.~\ref{fig:value_estimate_comparison_oreo}.

\end{document}